\pdfoutput=1

\documentclass[11pt]{article}

\usepackage{acl}

\usepackage{times}
\usepackage{latexsym}

\usepackage[T1]{fontenc}

\usepackage[utf8]{inputenc}

\usepackage{microtype}

\usepackage{inconsolata}

\usepackage{graphicx}

\usepackage{lipsum}  
\usepackage{tabularx}
\usepackage{booktabs}

\usepackage{microtype}
\usepackage{type1cm}
\usepackage[american]{babel}
\usepackage{amssymb}
\usepackage{amsmath}
\usepackage{multirow}
\usepackage{xspace}
\DeclareMathAlphabet{\mathcalbf}{OMS}{pzc}{b}{n}
\usepackage{enumitem}
\usepackage{colortbl}

\RequirePackage{color}
\definecolor{darkgray}{gray}{0.40}
\definecolor{mediumgray}{gray}{0.60}
\definecolor{lightgray}{gray}{0.95}
\definecolor{ultralightgray}{gray}{0.98}
\definecolor{forestgreen}{rgb}{0.133, 0.545, 0.133}
\definecolor{orange}{rgb}{1, 0.86, 0.74}
\definecolor{lightergreen}{rgb}{0.95, 1, 0.88}

\usepackage{graphicx}
\DeclareGraphicsExtensions{.pdf,.ai,.jpg,.png}
\setkeys{Gin}{pagebox=artbox}
\graphicspath{{.}}

\newcommand{\bsfigure}[3][]{%
    \begin{figure}[t]
        \centering
        \includegraphics[#1]{#2}
        \caption{#3}\label{#2}%
    \end{figure}
}

\RequirePackage{type1cm}
\RequirePackage{color}
\RequirePackage{soul}
\setstcolor{blue}
\definecolor{violet}{rgb}{0.5,0.0,0.5}

\newsavebox\bscombox
\newcommand{\bscom}[3][]{%
    \sbox{\bscombox}{\fontsize{8}{9}\selectfont#1#2#3}
    \noindent
    \st{#2}{\selectfont
        \color{blue}#3\ifx\\#1\\\else{\fontsize{8}{9}\selectfont\color{violet}[#1]}\fi
    }
}

\newcommand{\un}[1]{\underline{#1}}

\usepackage{tikz}

\newcommand{\bspubtag}{
	\vspace*{-18.6cm}\hspace*{-0.5cm}
	{\fontsize{6}{8}\selectfont%
		\renewcommand{\arraystretch}{0.9}
		\begin{tabular}{l}
			Accepted to Findings of the Association for Computational Linguistics: EMNLP 2026
		\end{tabular}
}}

\begin{document}

\title{Generating Constructive Feedback on Stories via Reinforcement Learning}




\author{
	Maja Stahl \\
	Leibniz University Hannover \\
	\texttt{m.stahl@ai.uni-hannover.de} \And
	Timon Ziegenbein \\
	Leibniz University Hannover \\
	\texttt{t.ziegenbein@ai.uni-hannover.de} \AND
	Henning Wachsmuth \\
	Leibniz University Hannover, L3S Research Center \\
	\texttt{h.wachsmuth@ai.uni-hannover.de}
}




\date{}

\maketitle

\begin{abstract}

Constructive feedback is crucial for creative writers to refine their storytelling abilities. Since receiving feedback from human experts is often costly and time-intensive, large language models (LLMs) offer a scalable and efficient alternative as automatic writing assistants. Despite their potential, research indicates that LLM-generated feedback is often generic, lacks actionability, and fails to identify which writing issue is most critical. To address these limitations, we present a reinforcement learning approach that steers LLMs to generate constructive feedback without the need for ground-truth feedback. We train our model using group relative policy optimization (GRPO) with a novel multi-component reward function aiming at constructiveness: it prioritizes feedback that is uniquely tailored to the story, helps to improve story quality, and addresses the most critical writing issue. In automatic and human evaluation across three story corpora, our approach outperforms state-of-the-art LLMs (including Gemini) and competitive baselines. We find that providing actionable suggestions is the main driver of feedback constructiveness.

\end{abstract}

\bspubtag
\vspace*{17.6cm}\hspace*{0.5cm}

\bsfigure{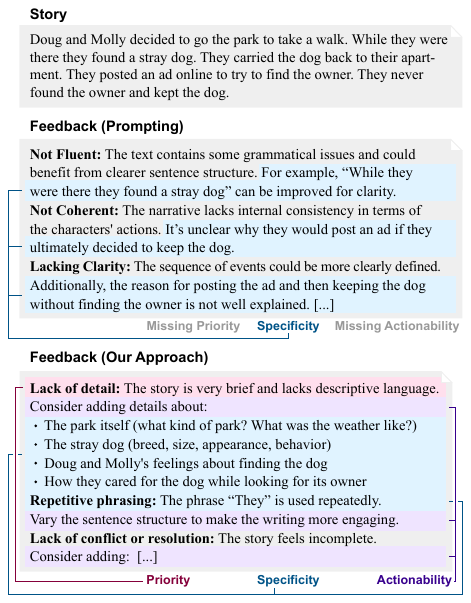}{Example story from \citet{rashkin-etal-2025-help} with two generated feedback texts, highlighting the criteria of constructiveness our approach improves compared to standard LLM prompting: (a) \emph{priority} in addressing the most critical issue, (b) \emph{specificity} to the given story, and (c) \emph{actionability} via concrete revision instructions.}

\section{Introduction}

Receiving feedback is an important aid in the writing process, helping authors improve their current text while refining their skills over time \cite{wolter-1975-effect,fatima-etal-2024-role}. Recent studies suggest that a considerate use of AI-based writing assistants may enhance writing performance for high school students \cite{ekizoglu-etal-2025-role} and language learners \cite{liu-etal-2024-integrating,mekheimer-2025-generative}, and also increase the efficiency and quality of academic writing \cite{khalifa-etal-2024-using}.

Within the creative writing domain specifically, research has explored NLP systems as writing assistants, ranging from fully-automated story generation \cite{fan-etal-2018-hierarchical,alhussain-azmi-2021-automatic} to interactive story writing tools that assist during the drafting process \cite{clark-etal-2018-creative, lee-etal-2022-coauthor,huot-etal-2025-agents}. Alternatively, prompting large language models (LLMs) can provide writing feedback on finished texts, but LLMs frequently miss critical issues \cite{rashkin-etal-2025-help} or hallucinate non-existent strengths and errors \cite{behzad-etal-2024-leaf}. Furthermore, without explicit quality validation, LLM-generated feedback often lacks both specific references to the text and actionable guidance \cite{chu-etal-2026-feedeval}. 

To address these limitations, \citet{nair-etal-2024-closing} and \citet{wang-liu-2025-real} introduce ``revision-as-reward'' methods that align feedback generation by explicitly modeling student revision behavior. However, these methods require simulating student behavior and rely on ground-truth revision data that is not publicly available. Consequently, standard LLM prompting remains the default, despite its persistent struggle to generate \emph{constructive} feedback \cite{rashkin-etal-2025-help}: feedback that is \emph{prioritized} (identifying the biggest problem), \emph{specific} (tailored to the story), and \emph{actionable} (providing clear, implementable suggestions).\,

In this paper, we study how to align LLMs to generate constructive writing feedback for creative stories  without the need for ground-truth revisions. To this end, we propose a reinforcement learning approach using group relative policy optimization (GRPO) \cite{shao-etal-2024-deepseekmath} with a novel multi-component reward function targeting the three core quality dimensions of constructive feedback outlined above. Figure~\ref{feedback-example} contrasts such constructive feedback to feedback that lacks priority (focusing on secondary problems) and actionability (missing concrete revision instructions). 

We systematically evaluate the need to optimize the three criteria to achieve constructive feedback, both with automatic metrics on three existing story corpora and in a human evaluation study. Our results demonstrate that our GRPO approach outperforms strong baselines, including Gemini 2.5 Flash. Moreover, they highlight that actionability is the primary driver of constructiveness, leading to feedback preferred by both LLM and human judges.

Altogether, this paper's main contributions are:\footnote{Our experiment code and data are available at \url{https://github.com/webis-de/EMNLP-26}.}
\begin{itemize}
    \setlength\itemsep{-0pt}
    \item A novel multi-component reward function employing scoring models that quantify feedback quality across three theory-informed criteria.
    \item The first reinforcement learning approach to generating constructive writing feedback without requiring revision data.
    \item Empirical evidence that our reward-based approach achieves a state-of-the-art alignment that robustly transfers across story corpora without corpus-specific human annotations.
\end{itemize}

\section{Related Work}

Computational writing assistance traditionally focused on specialized subtasks such as automated essay scoring \cite{ke-etal-2019-survey}, grammatical error detection \cite{gupta-2014-grammatical,bell-etal-2019-context}, and grammatical error correction \cite{felice-yuan-2014-generating,bryant-etal-2023-grammatical}. While these approaches effectively identify or fix surface-level errors, they often lack the explanatory depth needed for writing development. To address this, \citet{nagata-2019-toward} introduced the task of feedback comment generation, providing natural language explanations of \emph{why} specific sentences are erroneous, and \citet{stahl-wachsmuth-2023-identifying} show that explicitly modeling distinct feedback types in a multi-task setup improves such comment generation. Similarly, \citet{liu-etal-2024-geef} use encoder-decoder models to evaluate fluency and coherence for generating comments on essays, while \citet{han-etal-2019-level} propose revision suggestions to improve sentence-level proficiency. Whereas these prior works primarily target assessment or local edits, we focus on generating holistic, open-ended feedback to guide full-text revision.

With the rise of instruction-tuned large language models (LLMs), research has increasingly used prompting-based methods for feedback generation. \citet{stahl-etal-2024-exploring} investigate various prompting strategies for essay feedback, while \citet{behzad-etal-2024-leaf} utilize retrieval-augmented generation. Despite these advancements, studies reveal that LLM-generated feedback often misses critical writing issues \cite{rashkin-etal-2025-help} or hallucinates non-existent strengths and errors \cite{behzad-etal-2024-leaf}. Furthermore, without explicit quality validation, such feedback often lacks specific references to the text and actionable guidance \cite{chu-etal-2026-feedeval}. Our work aims to address these issues by aligning LLMs to generate constructive writing feedback.

To assist creative writing, some approaches generate stories automatically \cite{fan-etal-2018-hierarchical,alhussain-azmi-2021-automatic} or fine-tune models to suggest story continuations \cite{akoury-etal-2020-storium}. Others foster collaborative human-machine interaction, through a text editor where users collaborate with a prompted LLM \citep{yuan-etal-2022-wordcraft}, or through a multi-agent framework for incremental story building \citep{bae-kim-2024-collective}. While these systems effectively support the drafting phase, our work focuses on the post-draft feedback stage, which until now often relies on prompting. Specifically assessing this stage, \citet{rashkin-etal-2025-help} established an evaluation framework for LLM-generated story feedback using synthetic corruptions. They evaluate feedback across four criteria: whether it (i) is valid and story-specific, (ii) would improve the story if followed, (iii) identifies the primary writing issue, and (iv) provides appropriate positive reinforcement. In our work, we adopt the corpus-agnostic dimensions (i)--(iii) and refer to them as \emph{specificity}, \emph{actionability}, and \emph{priority}, respectively. Criterion (iv) can only be evaluated on their specific corpus and is therefore omitted here to enable robust evaluation across diverse corpora.

Since traditional overlap metrics like ROUGE and BLEU are ill-suited for capturing the pedagogical quality of such feedback, \citet{chu-etal-2026-feedeval} propose an LLM-based framework that evaluates and filters feedback along similar dimensions (specificity, helpfulness, and validity), leading to better correlation with expert human judgments. While our feedback criteria partially overlap with those of \citet{chu-etal-2026-feedeval}, we leverage these dimensions as reward signals to optimize the underlying generation policy via reinforcement learning, in contrast to post-hoc evaluation and filtering.

Beyond prompting, recent work explored direct preference optimization (DPO) \cite{rafailov-etal-2024-direct} and reinforcement learning (RL) for feedback generation. \citet{nair-etal-2024-closing} utilized DPO to learn from student revision performance as estimated by a simulator. Similarly, \citet{wang-liu-2025-real} optimized feedback via proximal policy optimization \cite{schulman-etal-2017-proximal}, balancing specificity and actionability against revision outcomes. In contrast, our approach evaluates the feedback itself rather than its downstream effects. We utilize group relative policy optimization (GRPO) \cite{shao-etal-2024-deepseekmath} to align the generation policy directly with human judgments on pedagogical quality. This eliminates the need for revision data and simulators during training, while simultaneously avoiding the inference-time latency of multi-agent refinement loops. To our knowledge, we are the first to propose an RL-based framework for writing feedback that operates without reference revisions. Specifically, we operationalize criteria from prior research into a novel multi-component reward function focused on three core dimensions: \emph{priority}, \emph{specificity}, and \emph{actionability} \cite{rashkin-etal-2025-help}.

\bsfigure{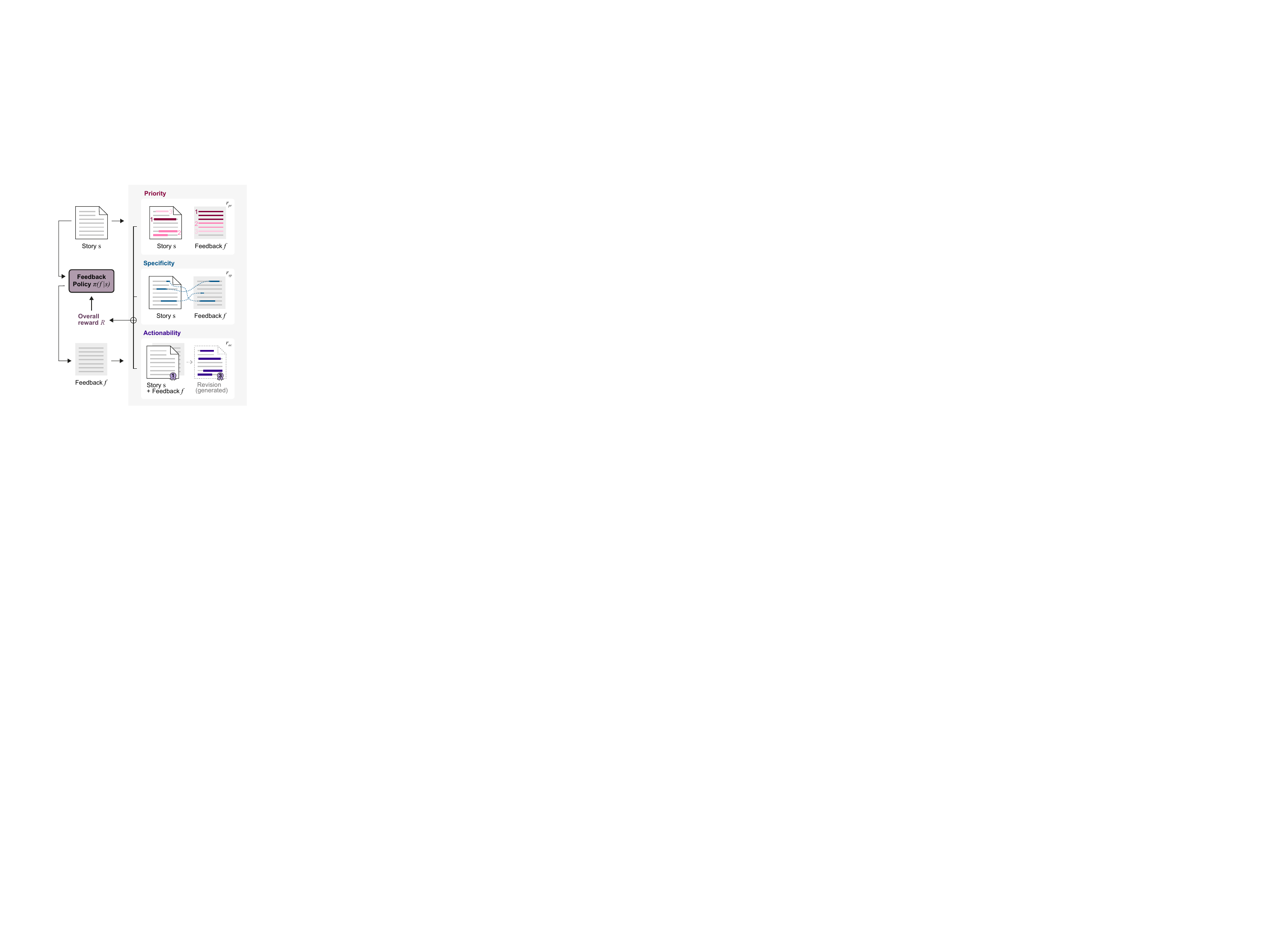}{Our reinforcement learning approach to story feedback generation, including the policy $\pi(f|s)$ for generating feedback $f$ on a story $s$, and the multi-component reward function $R(s,f)$ based on scoring models for priority, specificity, and actionability.}

\section{Approach}
\label{sec:approach}

This section presents our approach to generating constructive writing feedback for creative stories using reinforcement learning (RL). We employ a large language model (LLM) as a policy $\pi$, trained with group relative policy optimization (GRPO) \citep{shao-etal-2024-deepseekmath}, to generate natural language feedback for a given story. The core idea of our approach is to eliminate the need for ground-truth feedback or revision data during training by, instead, aligning the policy with a reward function derived from human annotations of feedback quality. Figure~\ref{approach-vertical-new2} gives an overview of the approach.

\subsection{Problem Definition}

Given a creative story $s$, our goal is to generate constructive textual feedback~$f$. Following
\citet{rashkin-etal-2025-help}, we consider feedback to be constructive if it satisfies the following three criteria:%
\footnote{These criteria map directly to \emph{error detection}, \emph{specificity}, and \emph{correctness} from \citet{rashkin-etal-2025-help}, but we adapted the terms to align with educational literature \cite{reigstad-mcandrew-1984-training,wiggins-2012-keys}. While \citet{rashkin-etal-2025-help} considers further criteria, only the given ones are corpus-agnostic and explicitly relate to constructiveness.}%
\begin{itemize}	
	\setlength\itemsep{0pt}
	\item \emph{Priority:} The feedback identifies and addresses the story's most critical writing issue.
	\item \emph{Specificity:} The feedback is tailored to the specific story and not applicable to other stories.
	\item \emph{Actionability:} Implementing the suggestions of the feedback improves the story's quality.
\end{itemize}

We hypothesize that explicitly optimizing for these criteria will align the LLM to generate highly constructive feedback. Crucially, by isolating these dimensions within our reward formulation, our approach allows us to systematically evaluate whether the three criteria all need to be optimized for in order to achieve constructive feedback.

\subsection{Reinforcement Learning}

Our RL policy, denoted as $\pi_\theta(f|s)$, is an LLM parameterized by $\theta$ that is instructed to generate constructive feedback~$f$ for a story~$s$. We use GRPO \cite{shao-etal-2024-deepseekmath}, which optimizes the policy by sampling a group of $k$ candidate feedback texts $\{f_1, f_2, \dots, f_k\}$ for each~$s$. This eliminates the need for a separate critic model by computing a relative advantage $A_i$ for each output based on the group's rewards $\{R(s, f_1), R(s, f_2), \dots, R(s, f_k)\}$:
\begin{eqnarray*}
	A_i &:=& \frac{R(s, f_i) - \text{mean}_{1 \leq j \leq k}(\{R(s, f_j)\})}{\text{std}_{1 \leq j \leq k}(\{R(s, f_j)\})}
\end{eqnarray*}

The objective is to find LLM parameters $\theta^*$ that maximize the expected reward:
\begin{eqnarray*}
	\theta^* &:=& \arg\max_\theta \mathbb{E}_{f \sim \pi_\theta(f|s)} [R(s, f)]
\end{eqnarray*}

\paragraph{Reward Components}

To quantify the feedback quality criteria defined above, we employ three separate scoring functions: $r_{pr}$, $r_{sp}$, and $r_{ac}$. During the RL process, they act as proxy reward functions that evaluate the generated feedback $f$ in the context of the story $s$. $r_{pr}(s, f)$ evaluates how well the biggest writing issue is addressed, $r_{sp}(s, f)$ assesses how well the feedback is tailored to the story and $r_{ac}(s, f)$ measures the story improvement potential of the feedback, each 
bounded within $[0,1]$:
\begin{eqnarray*}
	r_{c}: & \forall s \forall f: (s,f) \mapsto [0,1],   & c \in \{pr, sp, ac\}
\end{eqnarray*}

\paragraph{Overall Reward}

The overall reward $R(s, f)$ for a generated feedback text $f$ is defined as the arithmetic mean of the three components:
\begin{eqnarray*}
	R(s, f) & := &  \frac{1}{|C|} \sum_{c \in C} r_{c}(s, f), \;\;C = \{pr, sp, ac\}
\end{eqnarray*}

This ensures that the policy~$\pi$ treats each quality criterion with equal importance during the alignment process. Note that, while the scoring functions may have to be trained on annotated data, $\pi$ can be trained on any story corpus, without the need for corresponding feedback or quality labels.

\section{Data}
\label{sec:data}

The presented RL approach requires (a)~stories to align and evaluate its GRPO model, and (b)~quality-annotated feedback to train its reward models.

\subsection{Story Corpora}

We align and evaluate our approach on the following three corpora with human-written stories.

\paragraph{StoryFeedback {\rm \cite{rashkin-etal-2025-help}}} 

 contains 326 unique stories sampled from the ROCStories corpus \cite{mostafazadeh-etal-2016-corpus} and from three BIG-bench sources \cite{ghosh-srivastava-2022-epic, srivastava-etal-2023-beyond}. Each story was synthetically corrupted with one of three methods: \emph{backtranslation} to degrade grammar and tone, \emph{sentence swapping} to disrupt chronological order of events, and \emph{sentence deletion} to remove 
 context. Including the original stories, the corpus contains 1,304 stories, averaging 62 words \cite{rashkin-etal-2025-help}.

\paragraph{Storal {\rm \cite{guan-etal-2022-corpus} }} 

comprises 1,779 English stories designed to teach moral values. The stories were taken from the web and have an average length of 302 words.

\paragraph{WritingPrompts {\rm \cite{fan-etal-2018-hierarchical}}} 

contains 300,000 diverse fictional stories written in response to writing prompts from the \emph{WritingPrompts} subreddit. The stories are on average 735 words long.

\subsection{Quality-Annotated Feedback}

For a subset of the StoryFeedback corpus, \citet{rashkin-etal-2025-help} generated writing feedback using eight different LLMs and four prompting setups. Human evaluators then assessed feedback quality based on four questions: 
(a)~\emph{Encouragement:} Is the feedback giving positive comments appropriately? 
(b)~\emph{Priority}: Does it identify the biggest problem in the story? 
(c)~\emph{Specificity:} Is it valid feedback that's specific to the story? 
And (d)~\emph{Actionability:}~Would following the feedback make the story better? 
As defined in Section~\ref{sec:approach}, we utilize the annotations for the latter three dimensions to train our reward models, omitting the corpus-specific encouragement criterion. 
This resulted in 1,920 quality-annotated feedback texts, evenly distributed across generative models, prompting setups, and story versions.

\section{Experiments}
\label{sec:experiments}

This section describes the experiments that we conducted to evaluate the constructiveness of \nopagebreak{generated story} feedback. We first describe the training of our three reward models, which serve as the optimization objectives for quantifying the priority, specificity, and actionability of a feedback $f$ in the context of a story $s$. Then, we outline the experimental setup used to assess the effectiveness of our approach across diverse story corpora.

\subsection{Reward Models}

\newcolumntype{Y}{>{\raggedleft\arraybackslash}X}

\begin{table}[t]
	\centering
\small
	\setlength{\tabcolsep}{1pt} 
	\begin{tabularx}{\columnwidth}{l p{1pt} Y Y p{1pt} Y Y p{1pt} Y Y}
		\toprule
		\textbf{Model} && \multicolumn{2}{c}{\textbf{Priority}} && \multicolumn{2}{c}{\textbf{Specificity}} && \multicolumn{2}{c}{\textbf{Actionability}} \\
		\cmidrule{3-4} \cmidrule{6-7} \cmidrule{9-10} 
		&& \textbf{mse} $\downarrow$ & \textbf{pcc}  $\uparrow$ && \textbf{mse} $\downarrow$ & \textbf{pcc} $\uparrow$ && \textbf{mse} $\downarrow$ & \textbf{pcc} $\uparrow$ \\
		\midrule
		Random baseline     && .197 & .100 && .332 & -.096 && .213 & .008 \\
		Mean baseline       && .128 & .075 && .120 & .036  && .100 & .031 \\
		Reward model && \textbf{.092} & \textbf{.537} && \textbf{.031} & \textbf{.865} && \textbf{.058} & \textbf{.651} \\
		\bottomrule
	\end{tabularx}
	\caption{Reward model performance measured by mean squared error (\emph{mse}) and Pearson correlation coefficient (\emph{pcc}) for \emph{priority}, \emph{specificity}, and \emph{actionability}.}
	\label{tab:feedback-quality}
\end{table}

To provide the scalar signals required for policy alignment via GRPO, we develop three conceptually identical but independent reward models $r_{pr}(s, f)$, $r_{sp}(s, f)$, and $r_{ac}(s, f)$, corresponding to our quality criteria: \emph{priority}, \emph{specificity}, and \emph{actionability}. Using ModernBERT-large \citep{warner-etal-2025-smarter} with a single-output regression head, each reward model predicts human quality ratings from the StoryFeedback corpus \cite{rashkin-etal-2025-help}, given a concatenated story~$s$ and feedback~$f$:

%
\begin{verbatim}
	"Story: {story}\n\nFeedback: {feedback}"
\end{verbatim}

We optimize the reward models by maximizing the Pearson correlation coefficient $pcc$ between the predicted scores and human labels via Optuna \citep{akiba-etal-2019-optuna} over 20 trials. This objective prioritizes the relative ranking of instances over absolute value precision, ensuring the model captures the correct preference order. Each model output is clipped to the $[0, 1]$ interval, ensuring that the reward models function as bounded quality scorers.\,\,\,\,

Table~\ref{tab:feedback-quality} compares the effectiveness of each reward model (\emph{ModernBERT}) to a \emph{random} baseline, which samples scores uniformly from the range of training scores, and a \emph{mean} baseline, which predicts the mean label from the training set for all instances. Our reward models clearly outperform both baselines across all dimensions, achieving the lowest mean squared error and strong Pearson correlations with human ratings. Notably, they effectively capture \emph{specificity} ($pcc = .865$), while also correlating meaningfully with ground-truth \emph{priority} ($pcc = .537$) and \emph{actionability} ($pcc = .651$).

To validate these reward models beyond their training distribution, we additionally correlate their predictions with the independent human ratings from our manual evaluation (Section~\ref{sec:manualeval}), finding consistent, moderate to medium high correlations (Appendix~\ref{app:calibration}). Notably, our regression-based reward models correlate more strongly with human judgments than an LLM-as-a-judge ensemble would (e.g., $pcc = .43$ vs.\ $.38$ for overall constructiveness), supporting our choice of dedicated reward models over prompting-based judges as the reinforcement learning reward signal.

\subsection{Feedback Generation Models}

We instantiate our approach using three instruction-tuned LLMs as initial policy $\pi_{ref}$: \emph{\mbox{Gemma-2-9b-it}} \cite{gemmateam-2024-gemma}, \emph{Llama-3.1-8B-Instruct} \cite{grattafiori-etal-2024-llama3} and \emph{Qwen2.5-7B-Instruct} \cite{qwenteam-2024-qwen}. The base LLMs and our GRPO-based approach are instructed using the two most effective prompts from \citet{rashkin-etal-2025-help}: 
%
\begin{itemize}
	\setlength{\itemsep}{0pt}
	\item \emph{Open}: ``Write a bulleted list of feedback for the following text: 
	
	\{\emph{story}\}	
	
	If the story has no problems, say ‘The text is perfect as-is’.''
	\item \emph{Targeted}: ``Write a bulleted list of feedback for the following text:
	
	\{\emph{story}\} 
	
	The feedback should state whether the text is:
	\begin{itemize}
		\itemsep0em 
		\item not fluent (e.g. typos, grammar issues)
		\item not coherent (lacking internally consistency or badly ordered)
		\item lacking clarity (e.g. text is wordy or ambiguous)
		\item needs a stylistic change (e.g. formality, tone, other writing preferences)
		\item needs additions (e.g. needs more information or updated information)
		\item needs something removed (e.g. get rid of something that doesn’t fit with the rest of the document)
	\end{itemize}
	If the story has no problems, say ‘The text is perfect as-is’.''	
\end{itemize}

\subsection{Constructiveness Optimization}

We train our policies $\pi_\theta$ (instantiated from $\pi_{ref}$) using reinforcement learning on the StoryFeedback, Storal, and WritingPrompts corpora (see Section~\ref{sec:data}). To maintain consistency across datasets, we randomly downsample instances from Storal and WritingPrompts to match the 908 training examples in StoryFeedback. We use DrGRPO for optimization \cite{liu-etal-2025-understanding}, which mitigates length bias and optimization drift via global token-level normalization compared to standard GRPO \cite{shao-etal-2024-deepseekmath}. For each input story, we sample a group of $k = 8$ completions and optimize against our multi-component reward based on our reward models. To ensure parameter efficiency, we employ QLoRA \cite{dettmers-etal-2023-qlora}, which utilizes 4-bit NormalFloat quantization for the base model, while applying LoRA adapters \cite{hu-etal-2022-lora} (rank 16, $\alpha=32$) to all linear projections. For more details on hyperparameter configurations, see Appendix~\ref{app:training-details}.

\begin{table*}
	\setlength{\tabcolsep}{2.8pt}
	\centering
	\small 
	\begin{tabular}{ll cccc >{\columncolor{gray!15}}r r cccc >{\columncolor{gray!15}}r r cccc >{\columncolor{gray!15}}r}
		\toprule
		\bf Model & \bf Approach & \multicolumn{5}{c}{\bf (a) StoryFeedback} && \multicolumn{5}{c}{\bf (b) Storal} && \multicolumn{5}{c}{\bf (c) WritingPrompts} \\
		\cmidrule{3-7} \cmidrule{9-13} \cmidrule{15-19}
		& & \bf Pr & \bf Sp & \bf Ac & \bf Avg & \bf Rank$\downarrow$ && \bf Pr & \bf Sp & \bf Ac & \bf Avg & \bf Rank$\downarrow$ && \bf Pr & \bf Sp & \bf Ac & \bf Avg & \bf Rank$\downarrow$ \\
		\midrule
		Gemma 	& Base$_{\text{targeted}}$    & .718 & .912 & .763 & .798 & 9.41\phantom{$^\ddag$} 						&& .688 & .862 & .727 & .759 & 10.46\phantom{$^\ddag$} 								&& .734 & .919 & .757 & .803 & 8.19\phantom{$^\ddag$} \\
		& GRPO$_{\text{targeted}}$    & \un{.952} & .993 & \un{.930} & \un{.958} & 7.33$^\ddag$  				&& \un{.874} & \un{.987} & \un{.860} & \un{.907} & 5.89$^\ddag$ 					&& \un{.907} & .974 & \un{.875} & \un{.919} & 4.32$^\ddag$ \\
		& Base$_{\text{open}}$        & .668 & .971 & .780 & .807 & 4.93\phantom{$^\ddag$}  					&& .655 & .951 & .733 & .780 & 5.26\phantom{$^\ddag$} 								&& .685 & .949 & .758 & .797 & 4.28\phantom{$^\ddag$} \\
		& GRPO$_{\text{open}}$        & .930 & \un{.998} & .914 & .947 & \un{\bf 4.76}\phantom{$^\ddag$} 		&& .825 & .984 & .823 & .877 & \un{\bf 4.72}\phantom{$^\ddag$} 						&& .845 & \un{.979} & .857 & .894 & \un{\bf 2.96}$^\ddag$ \\
		\addlinespace
		LLaMA 	& Base$_{\text{targeted}}$    & .803 & .966 & .790 & .853 & 8.02\phantom{$^\ddag$}  					&& .753 & .962 & .764 & .827 & 7.74\phantom{$^\ddag$} 								&& .790 & .954 & .772 & .839 & 8.59\phantom{$^\ddag$} \\
		& GRPO$_{\text{targeted}}$    & .945 & \un{\bf .999} & .931 & .958 & \un{6.07}$^\ddag$ 					&& \un{.881} & \un{\bf .997} & .883 & .920 & \un{5.33}$^\ddag$ 						&& \un{.929} & \un{\bf .990} & .890 & \un{.936} & 7.65$^\ddag$ \\
		& Base$_{\text{open}}$        & .820 & .976 & .852 & .883 & 6.54\phantom{$^\ddag$}  					&& .727 & .943 & .793 & .821 & 5.46\phantom{$^\ddag$} 								&& .765 & .940 & .776 & .827 & \un{7.35}\phantom{$^\ddag$} \\
		& GRPO$_{\text{open}}$        & \un{.951} & .999 & \un{.961} & \un{.970} & 6.76\phantom{$^\ddag$}  		&& .879 & .995 & \un{\bf .904} & \un{.926} & 6.62\phantom{$^\ddag$} 				&& .920 & .982 & \un{\bf .906} & .936 & 8.70\phantom{$^\ddag$} \\
		\addlinespace
		Qwen  	& Base$_{\text{targeted}}$    & .336 & .289 & .511 & .378 & 12.58\phantom{$^\ddag$} 					&& .499 & .581 & .609 & .563 & 12.00\phantom{$^\ddag$} 								&& .712 & .798 & .730 & .747 & 11.69\phantom{$^\ddag$} \\
		& GRPO$_{\text{targeted}}$    & \un{\bf .973} & \un{.997} & .943 & .971 & \un{7.40}$^\ddag$ 			&& \un{\bf .922} & \un{.991} & .881 & \un{\bf .931} & \un{6.28}$^\ddag$ 			&& \un{\bf .955} & \un{.983} & .897 & \un{\bf .945} & \un{7.99}$^\ddag$ \\
		& Base$_{\text{open}}$        & .508 & .535 & .640 & .561 & 12.36\phantom{$^\ddag$} 					&& .454 & .493 & .589 & .512 & 12.73\phantom{$^\ddag$} 								&& .690 & .820 & .734 & .748 & 11.94\phantom{$^\ddag$} \\
		& GRPO$_{\text{open}}$        & .957 & .992 & \un{\bf .966} & \un{\bf .972} & 7.79$^\ddag$  			&& .898 & .979 & \un{.903} & .927 & 7.43$^\ddag$ 									&& .916 & .968 & \un{.903} & .929 & 8.46$^\ddag$ \\
		\midrule
		Gemini & Base$_{\text{targeted}}$   & \un{.781} & \un{.944} & \un{.793} & \un{.840} & 5.77\phantom{$^\ddag$}  && \un{.751} & .936 & .755 & \un{.814} & \un{7.18}\phantom{$^\ddag$} && \un{.748} & \un{.913} & \un{.765} & \un{.809} & 6.79\phantom{$^\ddag$} \\
		& Base$_{\text{open}}$       & .728 & .927 & .789 & .815 & \un{5.27}\phantom{$^\ddag$} && .723 & \un{.946} & \un{.756} & .808 & 7.87\phantom{$^\ddag$} && .704 & .899 & .746 & .783 & \un{6.11}\phantom{$^\ddag$} \\
		\bottomrule
	\end{tabular}
	\caption{Automatic evaluation results of feedback generation on (a) StoryFeedback, (b) Storal, and (c) WritingPrompts: Mean priority (\emph{pr}), specificity (\emph{sp}), actionability (\emph{ac}), their average (\emph{avg}), and the Bradley-Terry average \emph{rank} computed from pairwise multi-LLM judgments (see Section~\ref{sec:autoeval}) for our approach \emph{GRPO}, the three instruction-tuned \emph{Base} LLMs, and Gemini using \emph{targeted} and \emph{open} prompts. Bold: best overall; underlined: best per model family; $\ddag$: significant \emph{rank} improvement over the \emph{Base} counterpart (Wilcoxon signed-rank test, $p < .05$).}
	\label{tab:main-results}
\end{table*}

\subsection{Automatic Evaluation}
\label{sec:autoeval}

We evaluate all models on the 200 StoryFeedback test instances and 200 downsampled Storal and WritingPrompts test instances using these metrics:

\paragraph{Feedback Quality Models}
We evaluate feedback quality across the three dimensions optimized during training: priority (\emph{Pr}), specificity (\emph{Sp}), and actionability (\emph{Ac}). This allows us to check how well LLM do in general on these metrics and to what extent our GRPO-based approach improves over this.
Using our trained reward models, we compute the scores per instance and report test set averages  (range $[0,1]$, higher scores are better).

\paragraph{Multi-LLM-as-a-Judge}

To provide a robust measure of relative feedback quality, we perform systematic pairwise comparisons using bigger versions of the trained LLMs: \emph{\mbox{Gemma-2-27b-it}}, \emph{\mbox{Llama-3.1-70b-Instruct}}, and \emph{\mbox{Qwen-2.5-72b-Instruct}}. For each story, the LLMs are presented with two feedback candidates and instructed to select the more constructive one (prompt in Appendix~\ref{app:prompts}).
To mitigate position bias, each LLM evaluates every pair twice using swapped presentation orders. We first resolve the preference per individual LLM, requiring consistency across both orderings, and then determine the final preference for the pair via majority vote among the three models. This process minimizes both position bias and the individual style preferences of any single LLM. To aggregate these discrete pairwise preferences into a global ranking, we apply Bradley-Terry aggregation \cite{bradley-terry-1952-rank} and report the resulting average \emph{Rank} per feedback model. Lower average ranks indicate higher global preference.

We validate the reliability of this ensemble by correlating its verdicts with our human evaluation, finding that the majority vote consistently achieves the highest correlation among all judge configurations (Appendix~\ref{app:calibration}), confirming the benefit of ensembling over any individual judge.

\subsection{Manual Evaluation}
\label{sec:manualeval}

To complement our automatic analysis, we conducted a human evaluation study with five native English speakers with experience in writing or editing.
The evaluators were recruited via Upwork at \$15 per hour. Each evaluator scored 600 feedback texts across four dimensions: \emph{priority}, \emph{specificity}, \emph{actionability}, and overall \emph{constructiveness}. The scores were provided on a 5-point Likert scale, with each instance rated by all five annotators. 

These instances were randomly sampled and evenly distributed across the three datasets and four approaches. The choice of approaches in this manual evaluation is informed by the results of our automatic evaluation (see Section~\ref{sec:manualresults}): we selected the untrained model \emph{Gemma+Base$_{open}$}, our full approach \emph{Gemma+GRPO$_{open}$}, the best-performing ablation variant \emph{Gemma+GRPO$_{open\_ac}$} and the strong closed-source model \emph{Gemini$_{open}$}. Further details on the study are in Appendix~\ref{app:annotation-guidelines}.

\section{Results}

We report the findings of our experiments across the three story corpora from Section~\ref{sec:data}, starting with our automatic evaluation and an ablation study of our reward components. To validate these automatic metrics, we further present the results of our manual evaluation, confirming the effectiveness of our approach in providing constructive feedback.

\subsection{Automatic Evaluation Results}

Table~\ref{tab:main-results} compares our \emph{GRPO} approach 
with the \emph{Base} models and the closed-source baseline \emph{Gemini} 2.5 Flash 
across the \emph{targeted} and \emph{open} prompt.

Across all datasets and models, our approach results in substantial improvements over instruction-tuned base models in terms of the mean priority, specificity, and actionability scores (e.g., for \emph{Gemma+Base}$_{\text{targeted}}$ on StoryFeedback, from .718 to .952 for \emph{priority} and from .912 to .993 for \emph{specificity}). This suggests that standard instruction tuning is insufficient to satisfy all three feedback quality criteria, and that direct optimization for constructiveness is effective. Accordingly, except for Llama with open prompt, our approach always receives a notably better Bradley-Terry rank than the \emph{Base} variant (e.g., 5.89 vs.\ 10.46 for Gemma with targeted prompt on Storal). The most drastic shift can be observed for the Qwen model. While \emph{Qwen+Base}$_{\text{targeted}}$ often struggles with identifying the most critical writing issues (e.g., priority score only .336 on StoryFeedback), \emph{Qwen+GRPO}$_{\text{targeted}}$ achieves the highest priority scores for all datasets. 

The three LLM judges exhibit a Krippendorff's $\alpha$ of 0.46, indicating moderate agreement, and demonstrate only minor model family-specific preference biases. These biases were further mitigated through our use of majority voting (see Appendix~\ref{app:llm-bias} for the full bias analysis).

Notably, our trained 7B--9B GRPO-based models consistently outperform Gemini 2.5 Flash. For example, on WritingPrompts, \emph{Gemma+GRPO$_{\text{open}}$} achieves a rank of 2.96, compared to Gemini's rank of 6.11. This suggests that for story feedback generation, a smaller, efficiently-tuned open-weights model can provide more constructive 
assistance than a state-of-the-art general-purpose API model.

We observe that, while the \emph{targeted} prompt (providing explicit writing issues to check) often leads to higher scores for priority and specificity, the \emph{open} setup typically achieves better (i.e., smaller) overall rank. This implies that providing models with too many constraints may lead to a fixed style of feedback that LLM judges find less constructive on average than a more open-ended feedback.

\paragraph{Ablation Study}

\begin{table*}
	\setlength{\tabcolsep}{3.5pt}
	\centering
	\small
	\begin{tabular}{l cccc >{\columncolor{gray!15}}r r cccc >{\columncolor{gray!15}}r r cccc >{\columncolor{gray!15}}r}
		\toprule
		\bf Approach & \multicolumn{5}{c}{\bf (a) StoryFeedback} && \multicolumn{5}{c}{\bf (b) Storal} && \multicolumn{5}{c}{\bf (c) WritingPrompts} \\
		\cmidrule{2-6} \cmidrule{8-12} \cmidrule{14-18}
		& \bf Pr  & \bf Sp   & \bf Ac   & \bf Avg   & \bf Rank$\downarrow$  && \bf Pr  & \bf Sp   & \bf Ac   & \bf Avg   & \bf Rank$\downarrow$ && \bf Pr  & \bf Sp   & \bf Ac   & \bf Avg   & \bf Rank$\downarrow$ \\
		\midrule
		Base$_{open}$      & .668      & .971     & .780      & .807      & 4.53      && .655      & .951      & .733      & .780      & 5.86 && .685      & .949      & .758      & .797      & 5.00 \\
		GRPO$_{open\_pr}$      & \bf .969  & \bf .998 & .901      & .956      & 5.94      && \bf .910  & .984      & .834      & \bf .909  & 5.47 && \bf .930  & .963      & .837      & .910      & 6.48 \\
		GRPO$_{open\_sp}$      & .724      & .991     & .808      & .841      & 4.18      && .692      & .982      & .753      & .809      & 4.01 && .709      & .976      & .789      & .825      & 4.26 \\
		GRPO$_{open\_ac}$      & .884      & .988     & \bf .950  & .941      & \bf 3.08  && .786      & .984      & \bf .860  & .877      & \bf 2.15 && .817  & \bf .979  & .882      & .893      & \bf 3.10 \\
		GRPO$_{open\_no\_pr}$  & .842      & .996     & .896      & .911      & 4.33      && .753      & .983      & .819      & .852      & 3.32 && .761      & \bf .979  & .835      & .858      & 3.37 \\
		GRPO$_{open\_no\_sp}$  & .951      & \bf .998 & .935      & \bf .961  & 4.22      && .859      & \bf .985  & .843      & .895      & 4.19 && .903      & .973      & \bf .897  & \bf .924  & 4.25 \\
		GRPO$_{open\_no\_ac}$  & .946      & \bf .998 & .883      & .942      & 5.71      && .850      & \bf .985  & .812      & .882      & 5.59 && .853      & .975      & .830      & .886      & 5.40 \\
		GRPO$_{open}$      & .930      & \bf .998 & .914      & .947      & 4.03      && .825      & .984      & .823      & .877      & 5.44 && .845      & \bf .979  & .857      & .894      & 4.13 \\
		\bottomrule	
	\end{tabular}
	\caption{Ablation study of reward signals in feedback generation using the Gemma model. We compare our full approach \emph{GRPO$_{open}$} against variants trained with a single reward signal ($open\_pr$/$open\_sp$/$open\_ac$), and all but one reward signal ($open\_no\_pr$/$open\_no\_sp$/$open\_no\_ac$). Metrics as in Table~\ref{tab:main-results}.}
	\label{tab:ablation}
\end{table*}

To study the contribution of each reward component to feedback quality, Table~\ref{tab:ablation} compares our full approach (\emph{Gemma+GRPO$_{open}$}) against variants trained on a single reward signal (\emph{GRPO$_{open\_pr/open\_sp/open\_ac}$}) 
or on all but one signal (\emph{GRPO$_{open\_no\_pr/open\_no\_sp/open\_no\_ac}$}).

We find that the different reward models generally have the expected impact on the corresponding reward scores: For priority and actionability, the variant trained solely on the respective criterion (\emph{GRPO$_{open\_pr/open\_ac}$}) achieves the highest score in that category across all datasets (e.g., priority score .969 for \emph{GRPO$_{open\_pr}$} on StoryFeedback). Specificity is already remarkably high for the untrained baseline (\emph{Base}$_{open}$), with all scores exceeding .948. This high level of specificity is maintained successfully across all ablation variants, regardless of the specific reward optimized.

Notably, actionability appears to be the dominant factor for the rank. \emph{GRPO$_{open\_ac}$} consistently achieves the best rank across all three corpora (e.g., 2.15 on Storal), outperforming the other variants. In contrast, while \emph{GRPO$_{open\_pr}$} maximizes the priority score (reaching .969 on StoryFeedback), it receives the poorest rank (5.94). Agreement for these ablation comparisons was lower ($\alpha = 0.34$), likely due to the subtle differences between variants that are all highly optimized. Conversely, excluding actionability (\emph{GRPO$_{open\_no\_ac}$}) 
sharply worsens rank compared to the full model, confirming its necessity. This suggests that identifying critical issues without offering clear improvement suggestions 
is perceived as less constructive by the LLM judges. 

Although the full \emph{GRPO$_{open}$} model does not achieve the top rank, it provides a balanced performance, maintaining high scores across all three constructiveness criteria while remaining competitive in the multi-LLM ranking, ensuring that the feedback is not only actionable but also tailored specifically and prioritized correctly.

\paragraph{Human-Made vs.\ Synthetical Errors}

To study the effectiveness of our models on human-made errors compared to synthetic corruptions, Figure~\ref{original-vs-corrupted} compares the performance of our models on original, human-authored stories (x-axis) with the performance on synthetic corruptions (backtranslation, sentence swapping, sentence deletion) from the StoryFeedback corpus (y-axis). 

We find that the \emph{open} prompt (lighter shades) tends to improve performance on original stories compared to the \emph{targeted} prompt (darker shades). We suspect this to be due to the reduced restriction, allowing for more flexibility to address any present issue. For Gemini, both prompts lead to a clear bias toward corrupted stories, underperforming on the original stories. This suggests that, while the model addresses the more obvious flaws in the corrupted stories, more nuanced human writing issues remain challenging. As seen before, our GRPO approach improves performance in general, with \emph{Gemma+GRPO$\_{open}$} notably surpassing both Gemini variants, even more prominently on the original stories (all scores in Appendix~\ref{app:corruption}).

\bsfigure{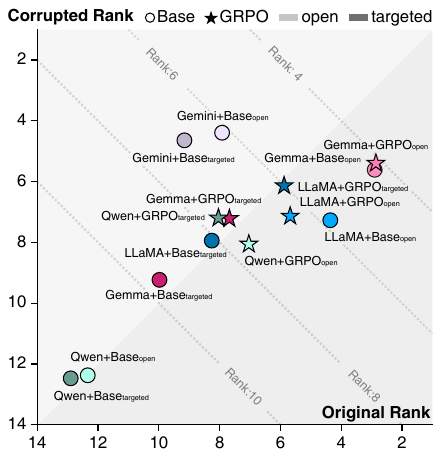}{Comparison of model performance on human-written stories (\emph{original rank}) versus synthetically-corrupted stories from the \emph{StoryFeedback} corpus (\emph{corrupted rank}). The dotted lines show the balanced \emph{rank}.}

\subsection{Manual Evaluation Results}
\label{sec:manualresults}

As outlined in Section~\ref{sec:manualeval}, five human annotators each evaluated 600 feedback instances 
to validate the automatic results (Krippendorff's $\alpha$ of 0.31). 

\begin{table}
	\setlength{\tabcolsep}{5pt}
	\centering
	\small 
	\begin{tabular}{ll cccc}
		\toprule
		\bf Approach && \bf Pr & \bf Sp & \bf Ac & \bf Co  \\
		\midrule
		Gemma+Base$_{\text{open}}$ 		&& 4.23 	& 4.81 		& 3.70 		& 3.55\phantom{$^\ddagger$}  	\\
		Gemma+GRPO$_{\text{open}}$ 		&& \un{4.24} & \un{4.85}	& \un{3.96} & \un{3.69}$^\ddagger$	\\
		Gemma+GRPO$_{\text{open\_ac}}$  && \bf 4.45 & \bf 4.90	& \bf 4.11	& \bf 4.01$^\ddagger$  \\
		\addlinespace
		Gemini+Base$_{\text{open}}$ 	&& 3.61 	& 4.40 		& 3.59 		& 3.19\phantom{$^\ddagger$}  	\\
		\bottomrule
	\end{tabular}
	\caption{Human evaluation of feedback in terms of priority (\emph{pr}), specificity (\emph{sp}), actionability (\emph{ac}), and constructiveness (\emph{co}). Our approach \emph{Gemma+GRPO$_{open\_ac}$} is best in all dimensions (bold), followed by the full approach \emph{Gemma+GRPO$_{open}$} (underlined). Both are significantly better ($\ddagger$) than \emph{Gemma+Base$_{open}$} (Wilcoxon signed-rank test with Bonferroni correction, $p < .05$).}
	\label{tab:manual-eval}
\end{table}

Table \ref{tab:manual-eval} shows that our GRPO approach optimized for actionability (\emph{Gemma+GRPO$_{open\_ac}$}) achieves the best rating across all dimensions (priority 4.45, specificity 4.90, actionability 4.11, constructiveness 4.01), followed by our full approach \emph{Gemma+GRPO$_{open}$}. Both significantly improve the overall constructiveness of feedback compared to \emph{Gemma+Base$_{open}$}, most prominently through increased actionability (+0.41 and +0.26). Consistent with the automatic evaluation, feedback generated by all \emph{Gemma} variants remains remarkably specific, with an average specificity rating above 4.81. 

\begin{table}
	\setlength{\tabcolsep}{5pt}
	\centering
	\small 
	\begin{tabular}{l cccc}
		\toprule
		\bf Criterium 			& \bf Pr & \bf Sp & \bf Ac & \bf Co \\
		\midrule
		Priority (Pr) 			&   		& \bf 0.313 & 0.246 	& 0.452 \\
		Specificity (Sp) 		& 0.313 	&   		& 0.214 	& 0.226 \\
		Actionability (Ac) 		& 0.246 	& 0.214 	&   		& \bf 0.556 \\
		Constructiveness (Co) 	& \bf 0.452 & 0.226 	& \bf 0.556	&   \\
		\bottomrule
	\end{tabular}
	\caption{Kendall's $\tau$ correlation between the feedback criteria rated in the human evaluation. The highest value per colunm is marked in bold.}
	\label{tab:quality-correlations}
\end{table}

Correlation analysis (Table~\ref{tab:quality-correlations}) reveals that actionability and priority are the strongest predictors of overall constructiveness (Kendall's $\tau$ of 0.556 and 0.452, respectively). This confirms that providing suggestions that are easy to implement and address a story's most critical issue is perceived as the most constructive form of feedback. Furthermore, the comparatively low-to-moderate correlations among the remaining criteria pairs ($\tau \le .31$) indicate that priority, specificity, and actionability capture distinct, non-redundant aspects of feedback quality. Notably, the larger \emph{Gemini} model underperforms in the key dimensions \emph{actionability} and \emph{priority}, resulting in lower overall constructiveness ratings. We conclude that our approach defines the state of the art in generating constructive story feedback.

\section{Conclusion}

So far, prompting is the default for generating writing feedback with LLMs. However, the resulting feedback often lacks specific, prioritized, and actionable guidance. To make LLM feedback more constructive, we present the first reinforcement learning approach using GRPO that explicitly optimizes for theory-informed feedback quality criteria without relying on ground-truth feedback or revisions. Guided by a novel multi-component reward function targeting the three core dimensions priority, specificity, and actionability, our approach produces significantly more constructive  feedback than state-of-the-art baseline models on three creative story corpora. Strikingly, it proves especially effective at addressing nuanced errors in original, human-authored stories. Our ablation studies and manual evaluations confirm that while all three quality criteria are important, actionability serves as the primary driver of perceived constructiveness. Ultimately, our work establishes a new state of the art for generating constructive feedback that is robust across different base models and corpora.
\section{Limitations}

While our approach successfully teaches LLMs to generate constructive writing feedback for creative stories, we see the following three limitations.

First, our experiments and the corresponding reward models focus exclusively on English creative stories. The transferability of our approach to other genres and languages remains to be investigated, primarily because high-quality feedback annotations for training reward models are currently unavailable for other settings. While it thus remains an open question as to how well the trained models generalize to other writing domains, such as academic or technical writing, we expect that the general process should apply to these as well.

Second, the assessment of feedback quality in creative writing is inherently subjective. Because creative writing relies heavily on the author's intent, what constitutes the most critical issue or a universally ``good'' suggestion can vary significantly. Our model inherently aligns with the consensus of the annotators who provided the data for the reward models. This might not perfectly match every individual author's stylistic vision or cultural context, but our approach could be simply retrained on respective other data, if available.

Third, our multi-component reward function focuses on priority, specificity, and actionability, but it lacks a dedicated ``validity'' or ``faithfulness'' component to enforce factual accuracy. However, our manual evaluation data suggests that severe quality problems are rare: Only 2.8\% of feedback generated by our approach received the lowest overall quality score (1/5) and 4.7\% the second-lowest (2/5). This is a significant and clear improvement over the baselines (3.9\% / 10.7\% for Gemma, and 10.7\% / 19.2\% for Gemini), highlighting that fully incorrect suggestions are not only the exception, but are actively reduced by our approach. Still, this analysis does not substitute for an explicit faithfulness constraint, and future research may integrate factuality criteria into the optimization process.

\section{Ethical Considerations}

While our approach is designed to support authors through constructive feedback, we cannot fully guarantee it may not occasionally produce feedback that misleads the author rather than guiding them in their writing skill development. However, we observed no such cases during inspection.

Furthermore, relying too heavily on such AI assistance poses educational risks. If an AI system consistently provides explicit, ready-to-use solutions, writers may passively accept these suggestions rather than thinking critically about how to resolve issues themselves. Over time, this over-reliance can substantially decrease the user's own learning effect, preventing them from independently developing their writing and editing skills.

Moreover, although our paper focuses on optimizing feedback quality, there is a risk of homogenizing creative expression. Because our reward models are aligned with the consensus of specific annotators, the system may implicitly enforce a normative, culturally specific standard of ``good'' writing. This could potentially suppress non-standard dialects, unconventional stylistic choices, or culturally distinct storytelling methods. We cannot fully prevent this bias, even though it is neither the intended use of our approach, nor is the approach optimized toward reducing creative diversity.

\section*{Acknowledgments}

The computational experiments in this paper have been supported by the Federal Ministry of Education and Research (BMBF), Germany, as part of the AI service center KISSKI (grant number 01IS22093C).

\bibliography{emnlp26-stories-lit}

\appendix





\section{Training Details}
\label{app:training-details}

\paragraph{Hyperparameters}Table \ref{tab:hyperparams} summarizes the configuration used for all GRPO training runs across the three model families (Gemma, LLaMA, and Qwen). We found that maintaining a consistent set of hyperparameters was sufficient for convergence across all three models and datasets, likely due to the stabilizing effect of the DrGRPO normalization.

\begin{table}
	\centering
	\small
	\begin{tabular}{lr}
		\toprule 
		\bf Hyperparameter & \bf Value \\
		\midrule
		Optimizer & Paged AdamW 8-bit \\
		Learning Rate & $5 \times 10^{-6}$ \\
		Learning Rate Scheduler & Cosine \\
		Warmup Ratio & 0.03 \\
		Weight Decay & 0.01 \\
		\midrule
		GRPO Group Size ($G$) & 8 \\
		KL Coefficient ($\beta$) & 0.002 \\
		DrGRPO Normalization & Token-level \\
		\midrule
		LoRA Rank ($r$) & 16 \\
		LoRA Alpha ($\alpha$) & 32 \\
		LoRA Target Layers & All Linear Projections \\
		Quantization & 4-bit NF4 \\
		\midrule
		Max Prompt Length & 1024 tokens \\
		Max Completion Length & 512 tokens \\
		Global Batch Size & 64 groups \\
		Gradient Accumulation & 16 steps \\
		\bottomrule
	\end{tabular}
	\caption{Hyperparameters used for GRPO training.}
	\label{tab:hyperparams}
\end{table}

\paragraph{Hardware and Environment}

All training was conducted on a cluster node equipped with 4$\times$ NVIDIA H100 (80GB) GPUs. We utilized \texttt{bf16} mixed precision for both training and reward model inference. The software environment was built on PyTorch 2.5 and the HuggingFace \texttt{accelerate} library. To ensure reproducibility, we set a global deterministic seed and isolated model caches for each run. Each full training run (3 epochs) took approximately 16 to 22 hours depending on the base model size.

\section{Multi-LLM-as-a-Judge Prompt}
\label{app:prompts}

\begin{quote}
	``You are evaluating two feedback texts (A and B) written for the same story. Assess which feedback is better for the story. For this, evaluate whether the feedback is specific to the story, whether following it would improve it, and whether it successfully addresses the story's biggest issue. Return only JSON with keys: winner (A|B|Tie), confidence (0..1), rationale.''
\end{quote}

\section{LLM Judge Bias Analysis}
\label{app:llm-bias}

Recent work found that LLMs have the tendency to favor their own outputs \cite{wang-etal-2024-large-language-models-fair}. We therefore investigate to what extent our judge models favor outputs generated by their own model family. To isolate the bias from the actual quality of the generated text, we calculate how much more a judge prefers its own family compared to how much other judges prefer that same family. For a judge model $J$ and a generating model $G$:

\begin{align*}
	\Delta_{self}(G) &= WR(J_{same}, G) \\
	&\quad - \text{avg}(WR(J_{others}, G))
\end{align*}

where $WR$ is the win rate (defined below). A positive $\Delta_{self}$ suggests a family-specific preference and a near zero $\Delta_{self}$ suggests the judge is objective and its preferences align with the general consensus of the other judge models. In line with  \citet{zheng-etal-2023-judging}, we calculate the win rate $WR$ as follows:

\begin{equation*}
	WR(J, G) = \frac{\#\text{Wins} + 0.5 \times \#\text{Ties}}{\#\text{Total Comparisons}}
\end{equation*}

Table~\ref{tab:judge-bias} presents the full relative preference matrix for all used judge and generating models. The diagonal elements correspond to the self-preference delta ($\Delta_{self}$) for each family. Gemma (27B) and LLaMA (72B) both exhibit a slight positive self-preference bias, with win rates increasing by +3.80\% and +2.63\% respectively when evaluating their smaller counterparts compared to the other judges. 
In contrast, Qwen (72B) shows the opposite effect ($\Delta_{self} = -2.78\%$), appearing more critical of its own family’s outputs than the other judges. 
However, the preference biases are rather small in our case and we use the majority voting over all judge models per instance to further mitigate such biases.

\begin{table}
	\setlength{\tabcolsep}{3pt}
	\centering
	\small 
	\begin{tabular}{l rrr}
		\toprule
		\bf Judge Model & \bf Gemma (9B) & \bf LLaMA (8B) & \bf Qwen (7B) \\
		\midrule
		Gemma (27B) & {\bf 3.80}\%  & -5.07\% 		& 1.26\% \\
		LLaMA (72B) & -4.15\% 		& {\bf 2.63}\%  & 1.52\% \\
		Qwen  (72B) & 0.35\% 		& 2.43\%  		& {\bf -2.78}\% \\
		\bottomrule
	\end{tabular}
	\caption{Relative preference matrix: Bold diagonal elements indicate the self-preference delta ($\Delta_{self}$), representing the deviation in win rate when a judge evaluates its own model family compared to the cross-family average. Positive values denote a relative preference, while negative values indicate a relative aversion.}
	\label{tab:judge-bias}
\end{table}


\section{Original vs Corrupted Stories}
\label{app:corruption}

\begin{table*}
	\setlength{\tabcolsep}{2pt}
	\centering
	\small 
	\begin{tabular}{ll crcc >{\columncolor{gray!15}}r r cccc >{\columncolor{gray!15}}r r cccc >{\columncolor{gray!15}}r}
		\toprule
		\bf Model & \bf Approach & \multicolumn{5}{c}{\bf (a) Original Stories} && \multicolumn{5}{c}{\bf (b) Corrupted Stories} && \multicolumn{5}{c}{\bf (c) All Stories} \\
		\cmidrule{3-7} \cmidrule{9-13} \cmidrule{15-19}
		& & \bf Pr & \bf Sp & \bf Ac & \bf Avg & \bf Rank$\downarrow$ && \bf Pr & \bf Sp & \bf Ac & \bf Avg & \bf Rank$\downarrow$ && \bf Pr & \bf Sp & \bf Ac & \bf Avg & \bf Rank$\downarrow$ \\
		\midrule                                                                                                                                                                                                                                                                  
		Gemma & Base$_{\text{targeted}}$    & .672          & .857          & .740            & .757           & 9.98          && .733          & .931          & .771           & .812          & 9.23          && .718          & .912          & .763          & .798          & 9.42          \\
		& GRPO$_{\text{targeted}}$          & \un{.938}     & .995          & \un{.922}       & \un{.952}      & 7.68          && \un{.954}     & .994          & \un{.927}      & \un{.959}     & 7.21          && \un{.950}     & .995          & \un{.926}     & \un{.958}     & 7.33          \\
		& Base$_{\text{open}}$   			& .648          & .966          & .770            & .795           & 2.90          && .675          & .973          & .783           & .810          & 5.61          && .668          & .971          & .780          & .807          & 4.93          \\
		& GRPO$_{\text{open}}$              & .918          & \un{.997}     & .908            & .941           & \un{\bf 2.86} && .934          & \un{.998}     & .916           & .949          & \un{5.39}     && .930          & \un{.998}     & .914          & .947          & \un{\bf 4.76} \\
		\addlinespace                                                                                                                                                                                                                                                                      
		LLaMA & Base$_{\text{targeted}}$    & .787          & .976          & .782            & .848           & 8.26          && .809          & .963          & .793           & .855          & 7.94          && .803          & .966          & .790          & .853          & 8.02          \\
		& GRPO$_{\text{targeted}}$          & .927          & \un{\bf 1.000}& .922            & .950           & 5.88          && .951          & \un{\bf .999} & .934           & .961          & \un{6.13}     && .945          & \un{\bf .999} & .931          & .958          & \un{6.07}     \\
		& Base$_{\text{open}}$   				& .779          & .972          & .840            & .863           & \un{4.36}     && .834          & .978          & .856           & .889          & 7.27          && .820          & .976          & .852          & .883          & 6.54          \\
		& GRPO$_{\text{open}}$              & \un{.937}     & .999          & \un{\bf .954}   & \un{\bf .963}  & 5.68          && \un{.956}     & \un{\bf .999} & \un{.963}      & \un{.973}     & 7.13          && \un{.951}     & \un{\bf .999} & \un{.961}     & \un{.970}     & 6.77          \\
		\addlinespace                                                                                                                                                                                                                                                                    
		Qwen & Base$_{\text{targeted}}$     & .276          & .229          & .475            & .327           & 12.90         && .356          & .308          & .523           & .396          & 12.47         && .336          & .289          & .511          & .378          & 12.58         \\
		& GRPO$_{\text{targeted}}$          & \un{\bf .959} & \un{.997}     & .932            & \un{\bf .963}  & \un{8.04}     && \un{\bf .978} & \un{.997}     & .947           & .974          & \un{7.19} 	 && \un{\bf .973} & \un{.997}     & .943          & .971          & \un{7.40} 	  \\
		& Base$_{\text{open}}$    				& .433          & .452          & .593            & .493           & 12.34         && .534          & .563          & .656           & .584          & 12.37         && .508          & .535          & .640          & .561          & 12.36         \\
		& GRPO$_{\text{open}}$              & .935          & .990          & \un{\bf .954}   & .960           & 7.04          && .965          & .993          & \un{\bf .969}  & \un{\bf .976} & 8.05          && .957          & .992          & \un{\bf .966} & \un{\bf .972} & 7.80          \\
		\midrule                                                                                                                                                                                                                                                                    
		Gemini & Base$_{\text{targeted}}$   & \un{.699}     & \un{.879}     & .747            & \un{.775}      & 9.16          && \un{.809}     & \un{.966}     & \un{.809}      & \un{.861}     & 4.64     	 && \un{.781}     & \un{.944}     & \un{.793}     & \un{.840}     & 5.77	      \\
		& Base$_{\text{open}}$  			& .671          & .868          & \un{.754}       & .764           & \un{7.92}	   && .747          & .946          & .800           & .831          & \un{\bf 4.39} && .728          & .927          & .789          & .815          & \un{5.27}	  \\
		\bottomrule
	\end{tabular}
	\caption{Automatic feedback generation results on (a) original, (b) corrupted (backtranslated, sentence swapped, sentence deleted), and (c) all stories from the StoryFeedback corpus. We compare our approach \emph{GRPO} against instruction-tuned \emph{Base} LLMs and Gemini using \emph{targeted} and \emph{open} setups: Mean priority (\emph{pr}), specificity (\emph{sp}), and actionability (\emph{ac}) scores and the Bradley-Terry average \emph{rank} computed from pairwise multi-LLM judgments (see Section~\ref{sec:autoeval}). Underlining marks the best value per model family; bold marks the best overall.}
	\label{tab:corrupted-results}
\end{table*}

Table~\ref{tab:corrupted-results} shows the feedback generation performance on (a) original, (b) corrupted (backtranslated, sentence swapped, sentence deleted), and (c) all stories from the StoryFeedback corpus.

\section{Calibration of Automatic Metrics}
\label{app:calibration}

To assess how well our automatic metrics align with human perception, and to what extent our reward signals overlap with our LLM-judge-based ranking, we compute Pearson correlation coefficients ($pcc$) between (a) our reward models and the human ratings from our manual evaluation, (b) our LLM judgements and the same human ratings, and (c) our reward models and our LLM judgements.

\paragraph{Reward Models and Human Judgments}

Table~\ref{tab:reward-human-corr} reports the correlation between each reward model's predictions and the corresponding human ratings from our manual evaluation (Section~\ref{sec:manualeval}). Each human rating criterium correlates most strongly with its matching reward model's outputs, and the full reward correlates with overall human-rated constructiveness at $pcc = .43$.

\begin{table}
	\setlength{\tabcolsep}{9pt}
	\centering
	\small
	\begin{tabular}{l cccc}
		\toprule
		\bf Human \textbackslash\ Reward & \bf Pr & \bf Sp & \bf Ac & \bf Co \\
		\midrule
		Priority (Pr)        & \bf .35 & .30      & .29      & .31 \\
		Specificity (Sp)     & .37     & \bf .53  & .38      & .47 \\
		Actionability (Ac)   & .29     & .32      & \bf .42  & .34 \\
		Constructiveness (Co)         & .34     & .28      & .30      & \bf .43 \\
		\bottomrule
	\end{tabular}
	\caption{Pearson correlation between reward model predictions (columns) and human ratings from the manual evaluation (rows). Bold marks each reward component's correlation with its matching human dimension.}
	\label{tab:reward-human-corr}
\end{table}

\paragraph{LLM Judgments and Human Judgments}
Table~\ref{tab:judge-human-corr} reports the correlation between each LLM judge's pairwise verdicts and the human ratings. The majority-vote ensemble achieves the highest correlation across all four dimensions, confirming that aggregating judges reduces individual judge noise relative to human perception.

\begin{table}
	\setlength{\tabcolsep}{2pt}
	\centering
	\small
	\begin{tabular}{l cccc}
		\toprule
		\bf Human \textbackslash\ Judge & \bf Llama & \bf Gemma & \bf Qwen & \bf Majority \\
		\midrule
		Priority (Pr)        & .32 & .32 & .31 & \bf .34 \\
		Specificity (Sp)     & .37 & .36 & .33 & \bf .38 \\
		Actionability (Ac)   & .22 & .21 & .20 & \bf .23 \\
		Constructiveness (Co)         & .36 & .35 & .34 & \bf .38 \\
		\bottomrule
	\end{tabular}
	\caption{Pearson correlation between LLM judge verdicts (columns) and human ratings from the manual evaluation (rows). The majority-vote ensemble correlates most strongly with human judgments in every dimension (bold).}
	\label{tab:judge-human-corr}
\end{table}

\paragraph{Reward Models and LLM Judgments}

Since our automatic evaluation in Section~\ref{sec:autoeval} reports both reward-model scores and LLM-judge rankings, we quantify their overlap in Table~\ref{tab:reward-judge-corr} to rule out that the two automatic metrics are trivially redundant. We find moderate correlations, indicating that the reward-based and judge-based metrics capture related but non-identical aspects of feedback quality.

\begin{table}
	\setlength{\tabcolsep}{9pt}
	\centering
	\small
	\begin{tabular}{l cccc}
		\toprule
		\bf Judge \textbackslash\ Reward & \bf Pr & \bf Sp & \bf Ac & \bf Co \\
		\midrule
		Llama 72B    & .28 & .43 & .33 & .37 \\
		Gemma 27B    & .34 & .45 & .35 & .42 \\
		Qwen 72B     & .15 & .35 & .25 & .27 \\
		Majority     & .27 & .43 & .33 & .37 \\
		\bottomrule
	\end{tabular}
	\caption{Pearson correlation between reward model predictions (columns) and LLM judge verdicts (rows).}
	\label{tab:reward-judge-corr}
\end{table}

\paragraph{Comparison to LLM-as-a-Judge as Reward}

Comparing the diagonal correlations in Table~\ref{tab:reward-human-corr} and the column \emph{Majority} of Table~\ref{tab:judge-human-corr} shows that our regression-based reward models correlate with human judgments consistently more strongly than the LLM-judge ensemble (e.g., $.53$ vs.\ $.38$ for specificity, $.42$ vs.\ $.23$ for actionability), despite the latter's substantially higher inference cost. This supports our design choice of lightweight regression-based reward models for the RL loop.




\section{Manual Evaluation Study}
\label{app:annotation-guidelines}

The human evaluators were given the following guidelines for rating the feedback texts:

\paragraph{Priority}

\emph{Definition:} Does the feedback identify and address the story's most critical writing issue? High-priority feedback targets the core problem of a story rather than focusing on surface-level details.

\emph{Note:} If the story is already of very high quality (or you believe it has no issues), a high score should be given to feedback that acknowledges the story's strength or provides high-level thematic polish.

\begin{description}
	\item[\textbf{1 (Not Prioritized):}] The feedback misses the point entirely, or invents a problem that doesn't exist.
	\item[\textbf{2 (Rather Not Prioritized):}] The feedback focuses on trivialities (e.g., typos, formatting) while ignoring significant narrative flaws.
	\item[\textbf{3 (Partly Prioritized):}] The feedback addresses a valid but secondary issue.
	\item[\textbf{4 (Rather Prioritized):}] The feedback addresses a major issue, though perhaps not the absolute top priority.
	\item[\textbf{5 (Fully Prioritized):}] The feedback perfectly identifies the single most critical issue in the story.
\end{description}

\paragraph{Specificity}
\emph{Definition:} Is the feedback tailored to this specific story, or is it a ``template'' that could apply to any story? Specific feedback demonstrates that the evaluator has read and understood the text, using evidence from the story to support its claims.

\begin{description}
	\item[\textbf{1 (Not Specific):}] Generic feedback with zero reference to the story content.
	\item[\textbf{2 (Rather Not Specific):}] Uses broad terms (e.g., ``the protagonist,'' ``the ending'') that could apply to many stories in this genre.
	\item[\textbf{3 (Partly Specific):}] Mentions the general topic of the story, but lacks deep engagement with the unique text.
	\item[\textbf{4 (Rather Specific):}] Mentions clear details of the story, though it might use some standard feedback phrasing.
	\item[\textbf{5 (Fully Specific):}] Uses character names, quotes specific lines, or references unique plot points. It is impossible to apply this feedback to a different story.
\end{description}

\paragraph{Actionability}
\emph{Definition:} Does the feedback provide a clear path for the author to improve the story? Actionable feedback goes beyond pointing out a problem. It provides a suggestion or a ``how-to'' for the fix, leaving the writer knowing what their next steps are.

\begin{description}
	\item[\textbf{1 (Not Actionable):}] Simply states a preference or critique without any hint of a solution.
	\item[\textbf{2 (Rather Not Actionable):}] Offers advice that is too abstract to implement.
	\item[\textbf{3 (Partly Actionable):}] Identifies a weakness, but the author has to do a lot of ``heavy lifting'' to figure out how to fix it.
	\item[\textbf{4 (Rather Actionable):}] Clearly points out what needs to change, making it easy for the author to know their next steps.
	\item[\textbf{5 (Fully Actionable):}] Provides a clear how-to or even a concrete suggestion/example of how to rewrite a section.
\end{description}

\paragraph{Constructiveness}
\emph{Definition:} This is an overall measure of the feedback's quality. It considers whether the feedback is helpful, encouraging, and balances the three previous criteria. Following highly constructive feedback would result in a significantly better story.

\begin{description}
	\item[\textbf{1 (Not Constructive):}] The feedback is confusing, factually incorrect about the story, or entirely generic.
	\item[\textbf{2 (Rather Not Constructive):}] The feedback is too generic to be useful or focuses on the wrong things, providing little value to the writer.
	\item[\textbf{3 (Partly Constructive):}] The feedback is ``fine'', meaning it is not wrong, but it is not particularly insightful or transformative either.
	\item[\textbf{4 (Rather Constructive):}] Very helpful feedback that would definitely lead to a better second draft, with only minor room for more detail.
	\item[\textbf{5 (Fully Constructive):}] The ``Gold Standard.'' It is specific, hits the most important point, and gives the author a clear path forward.
\end{description}

Figure~\ref{fig:interface} shows the annotation interface. 

\onecolumn
\clearpage

\begingroup
\noindent
\centering
\includegraphics[width=1\textwidth]{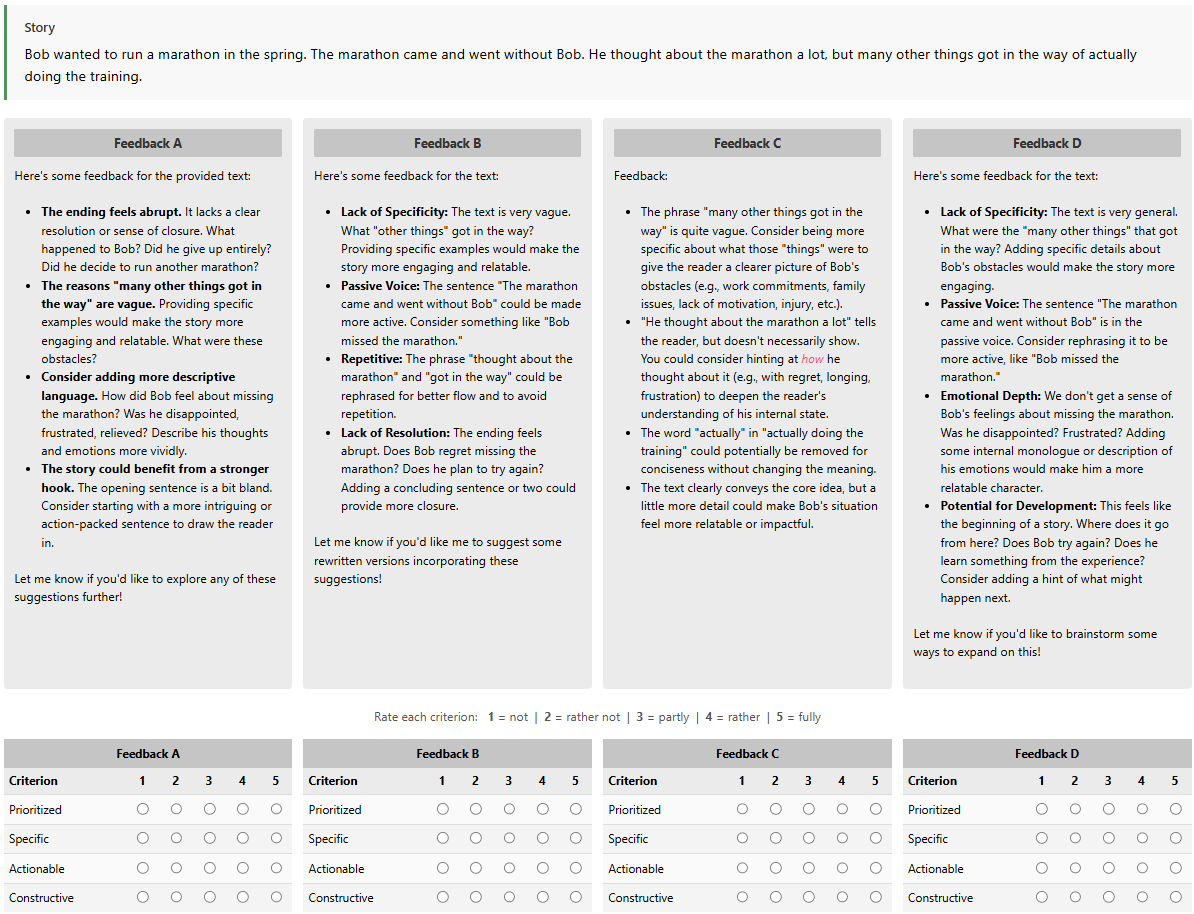}
\captionof{figure}{Annotation interface for evaluating feedback texts using a 5-point Likert scale across four quality dimensions (priority, specificity, actionablity and  constructiveness).}
\label{fig:interface}
\endgroup

\twocolumn

\end{document}